\documentclass{article}
\usepackage{ijcai21}

\usepackage{times}
\usepackage{soul}
\usepackage{bigstrut}
\usepackage{url}
\usepackage[hidelinks]{hyperref}
\usepackage[utf8]{inputenc}
\usepackage[small]{caption}
\usepackage{graphicx}
\usepackage{multirow}
\usepackage{mathtools}
\usepackage{tikz}
\usepackage{booktabs}
\usepackage{algorithm}
\usepackage{algorithmic}
\usepackage{amsmath}
\usepackage{amssymb}
\usepackage{amsfonts}
\usepackage{amsthm}

\def\checkmark{\tikz\fill[scale=0.4](0,.35) -- (.25,0) -- (1,.7) -- (.25,.15) -- cycle;} 

\title{Noise Doesn't Lie: Towards Universal Detection of Deep Inpainting}
\author{
Ang Li$^1$
\and
Qiuhong Ke$^{1*}$\and
Xingjun Ma$^2$\footnote{Corresponding authors.}\and
Haiqin Weng$^3$\and \\
Zhiyuan Zong$^3$\and 
Feng Xue$^3$\And
Rui Zhang$^4$
\affiliations
$^1$The University of Melbourne  \\
$^2$Deakin University\\
$^3$Ant Group\\
$^4$Tsinghua University
\emails
angl4@student.unimelb.edu.au,
qiuhong.ke@unimelb.edu.au,
daniel.ma@deakin.edu.au,
haiqin.wenghaiqin@antfin.com,
david.zzy@antgroup.com,
gkn1fexxx@gmail.com,
rayteam@yeah.net
}

\usepackage{xcolor}

\begin{document}

\maketitle

\begin{abstract}
Deep image inpainting aims to restore damaged or missing regions in an image with realistic contents.
% that seamlessly blend into the existing background.
% These techniques are known as deep inpainting and have been widely used to restore damaged or missing regions in an image.
While having a wide range of applications such as object removal and image recovery, deep inpainting techniques also have the risk of being manipulated for image forgery.
A promising countermeasure against such forgeries is deep inpainting detection, which aims to locate the inpainted regions in an image.
%In this paper, we propose a simple but very effective approach for deep inpainting detection.
In this paper, we make the first attempt towards \emph{universal} detection of deep inpainting,
where the detection network can generalize well when detecting \emph{different} deep inpainting methods.
%where the detection network trained on one \emph{single} type of training data can be applied to detect images tampered by \emph{different} deep inpainting methods.
%Specifically, we propose to use the noise pattern defined by the difference between an image and its reconstruction by an autoencoder to generate a universal training dataset, and further design a novel detection network that exploits the discriminative information contained in both the image and its noise pattern. 
To this end, we first propose a novel data generation approach to generate a \emph{universal} training dataset, which imitates the noise discrepancies exist in real versus inpainted image contents to train universal detectors.
We then design a Noise-Image Cross-fusion Network (NIX-Net) to effectively exploit the discriminative information contained in both the images and their noise patterns.
We empirically show, on multiple benchmark datasets, that our approach outperforms %can outperform
existing detection methods by a large margin and generalize well to unseen deep inpainting techniques. Our universal training dataset can also significantly boost the generalizability of existing detection methods.
%Moreover, our universal dataset can significantly boost the generalizability of existing methods, verifying the power of noise discrepancy for universal deep inpainting detection.

% Image inpainting with deep learning has recently achieved rapid development.
% Despite originally aimming at restoring damaged images, it can also be used for malicious intenions.
% Hence, detection on the modified regions of deep inpainting in an image is becoming an urgent need.
% However, current detection approaches have limited generalization ability and experience a significant performance drop when applied to unseen deep inpainting techniques.
% We observe that there exist intrinsic discrepancies on noise patterns between real image contents and contents generated by deep inpainting.
% This paper presents a universal deep inpainting detection approach using noise guidance. 
% We first propose a novel framework for universal training data generation without any knowledge of specific deep inpainting techniques.
% Then, we propose a novel detection network which utilizes the noise residual of input image as guidance in a multi-scale manner.
% Extensive experimental results have shown the effective generalization ability of the proposed approach.

\end{abstract}

\section{Introduction}
Image Inpainting is the process of restoring damaged or missing regions of a given image based on the information of the undamaged regions.
It has a wide range of real-world applications such as the restoration of damaged images and the removal of unwanted objects.
%A large amount of different empirical approaches have been proposed for the image inpainting task, which can be divided into two categories:  diffusion-based approaches \cite{935036,1238360} and patch-based approaches \cite{790383,wexler2007space,barnes2009patchmatch}.
%Both of the two groups lack the ability to yield unseen patterns or complex structures which are not available in the
So far, plenty of inpainting approaches have been proposed, among which generative adversarial networks (GANs) \cite{goodfellow2014generative} based deep inpainting techniques \cite{pathak2016context,iizuka2017globally,yu2018generative,li2019generative,Yu_2019_ICCV,Li_2020_CVPR} 
% based on
% generative adversarial networks (GANs) \cite{goodfellow2014generative} %based deep image inpainting techniques \cite{pathak2016context,Yang_2017_CVPR,iizuka2017globally,yu2018generative,liu2018image,Yu_2019_ICCV} 
have been demonstrated to be the most effective ones.
% With the rocketing development of deep learning, various deep image inpainting techniques \cite{pathak2016context,Yang_2017_CVPR,iizuka2017globally,yu2018generative,liu2018image,Yu_2019_ICCV} have sprouted out in recent years.
One distinguished advantage of deep inpainting models is the ability to adaptively predict semantic structures and produce super realistic and fine-detailed textures. 
%As such, they are flexible enough to generate visually plausible contents for a wide variety of scene images such as human faces, real-life objects and natural views.

\begin{figure}[t!]
\centering
\setlength{\tabcolsep}{0.05em}
{
\begin{tabular}{cccc}

\includegraphics[width=2.1cm]{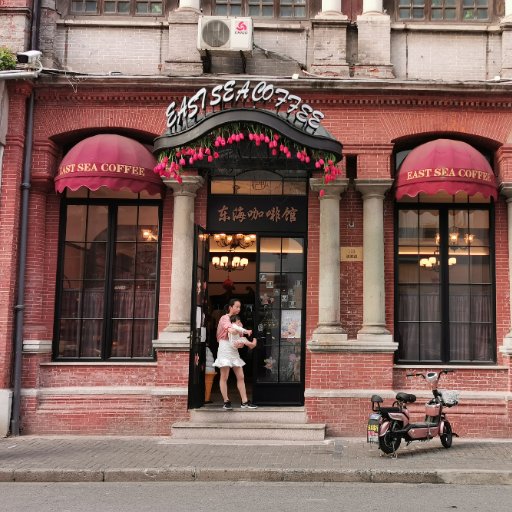}&
\includegraphics[width=2.1cm]{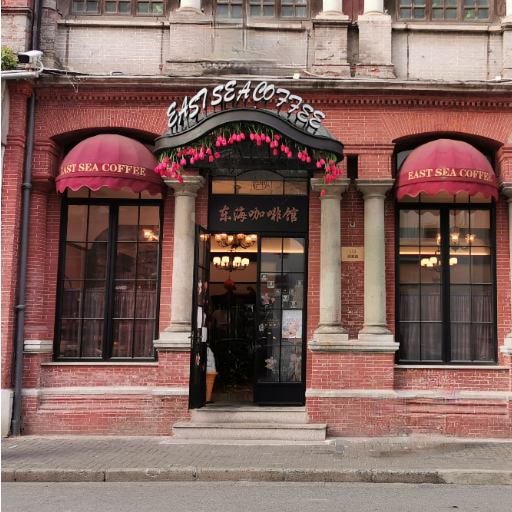}&
\includegraphics[width=2.1cm]{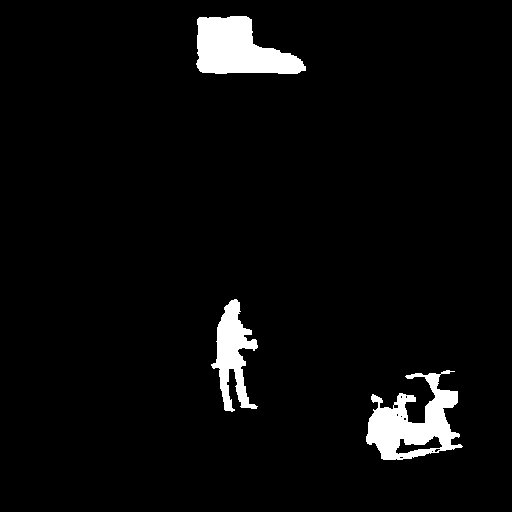}&
\includegraphics[width=2.1cm]{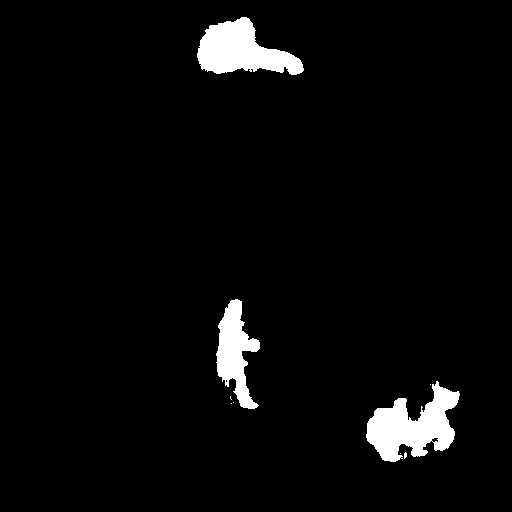}
\\

\scriptsize (a) Original image & \scriptsize(b) Inpainted image & \scriptsize(c) Inpainting mask & \scriptsize(d) Detected mask  \\

\end{tabular}
}
\caption{(a) The original image; (b) The inpainted image by the deep inpainting method in \protect\cite{Yu_2019_ICCV}; (c) The mask that defines the inpainting region; (d) The detected mask by our approach.}
\label{ijcai2021:exampleCompare} % I can do without the label too
\end{figure}
% \vspace{-0.1in}

However, like a two-edged sword cuts both ways, deep inpainting techniques come along with the risk of being manipulated for image forgery.
Due to the super realistic inpainting effects,  these techniques can be easily applied to replace the critical objects in an image with fake contents, and the tampered image may appear as photo-realistic as real images. 
% Even humans cannot easily differentiate those tampered images from the real ones.
% This makes it extremely hard to differentiate such modified images from real images even by human eyes.
Figure~\ref{ijcai2021:exampleCompare} (b) shows one such example crafted from the real image in Figure~\ref{ijcai2021:exampleCompare} (a) by a recent deep inpainting method \cite{Yu_2019_ICCV}. 
% Particularly with the state-of-the-art inpainting approaches, the tampered images can be as photo-realistic as real images.
% Even with careful inspection, human eyes may fail to find out the tampered regions.
% As illustrated in Figure~\ref{ijcai2021:exampleCompare}, three key objects in (a) are first removed with the mask in (c) and then filled by the inpainting approach \cite{Yu_2019_ICCV}, finally obtaining a visually plausible inpainted image in (b).
Inpainted images can potentially be used to create fake news, spread rumors on the internet or even fabricate false evidences.
It is thus imperative to develop detection algorithms to identify whether and more importantly \emph{where} an image has been modified by deep inpainting. Specifically, the goal of deep inpainting detection is to locate the exact inpainted regions in an image, as shown in Figure~\ref{ijcai2021:exampleCompare} (d).

% Forensics of empirical inpainting approaches are also studied \cite{li2017localization,zhu2018deep}. 
% However, these detection approaches are incompatible with deep learning based inpainting forensics, due to the intrinsic difference that deep inpainting directly predicts the missing regions with perceptually consistent contents while conventional manipulation approaches cannot.
% Although deepfake detection studies have shown impressive performance on deep face forensics, they have little in common with deep inpainting detection.
% Specifically, deepfake manipulations have complex post-processing operations such as color transfer and boundary blending, while deep inpainting does not have.

% The most outstanding challenge of deep inpainting detection is the generalizability to unseen deep inpainting techniques.
%While several deep inpainting detectors have been proposed  \cite{li2017localization,zhu2018deep}, they all transfer poorly to deep inpainting techniques. 

While Li et al. \shortcite{li2019localization} recently proposed the first method for deep inpainting detection, its effectiveness is restricted to the inpainting technique the detector was trained on and does not generalize well to other inpainting techniques.
However, in real-world scenarios, the exact techniques used to inpaint the images are often unknown.
% can be tampered by different deep inpainting approaches which are hard to anticipate.
%Although extremely challenging to achieve, high generalizability is arguably the most desired property of a good deep inpainting detector. 
%This is because, in real-world scenarios, it is almost impossible to predict the exact deep inpainting technique used. 
In this paper, we aim to address this generalization limitation %of generalization
and introduce a \emph{universal} detector that  works well even on unseen deep inpainting techniques.
% Currently, the most outstanding challenge of image inpainting detection is the generalization ability.
% In real-world situations, the model actually needs to detect an inpainted image without knowing the underlying specific deep inpainting approach. 
% Li et al. recently propose the first work \shortcite{li2019localization} on deep generative inpainting detection by training a ResNet-based \cite{he2016identity} model on images manipulated by a known deep inpainting approach \cite{iizuka2017globally}.
% However, it struggles with overfitting and hence its effectiveness is restricted to the inpainting approaches specifically trained with. 
% When dealing with images forged by unknown deep inpainting approaches, it is of paramount importance to develop a detection model with stable generalization capability.

\begin{figure}[t!]
\centering
\setlength{\tabcolsep}{0.2em}
{
\begin{tabular}{ccc}

\includegraphics[width=2.0cm]{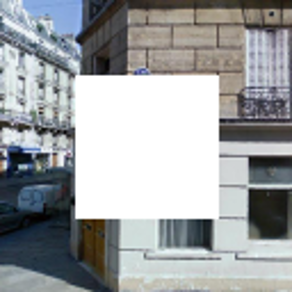}&
\includegraphics[width=2.0cm]{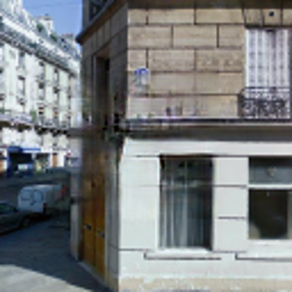}&
\includegraphics[width=2.0cm]{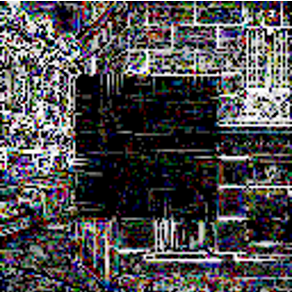}
\\

\scriptsize (a) Masked image & \scriptsize (b) Inpainted image  & \scriptsize (c) Noise residual  \\

\end{tabular}
}
\caption{Noise analysis on an inpainted image. (a) The image with the inpainting mask (the white region); (b) The inpainted image by the method in \protect\cite{Yu_2019_ICCV}; (c) The pixel-wise difference between (b) and a denoised version of (b), i.e., a noise residual pattern.}
\label{ijcai2021:noise} % I can do without the label too
\end{figure}

Our approach is motivated by one important yet so far overlooked common characteristic of all deep inpainting methods: the patterns of the noise exists in real and synthesized contents are different.
% may exist intrinsic noise discrepancies between the contents of 
% real images and the generated contents.
It has been shown that conventional image acquisition devices (e.g. camera sensors) leave distinctive noise patterns on each image~\cite{Farid2009}, however, GAN-generated contents do not have the same type of noise patterns~\cite{8695364}. As shown in Figure \ref{ijcai2021:noise},
given an inpainted image that contains both the device-captured real pixels and GAN-generated inpainted  pixels, %
%Figure~\ref{ijcai2021:noise} shows the  noise patterns  on  an inpainted image, where 
there are obvious discrepancies between inpainted region and non-inpainted region in the noise residual  (i.e., pixel-wise subtraction between the inpainted image and its denoised version by a denoising filter).
 %there is a distinctive noise-residual  (i.e., pixel-wise subtraction between the inpainted image and its denoised version by a denoising filter) pattern in the inpainted region compared with the surrounding region. %in the inpainted region can be observed. 
% Although hardly visible in the RGB image, it is obvious to notice the noise discrepancies between real contents  (the undamaged region) and generated contents (the inpainted region).
Motivated by this observation, we propose a universal deep inpainting detection framework with the following two important designs. First, we present a novel way of generating universal dataset for training universal detectors. 
The universal training dataset contains: 1) images with synthesized (using a pre-trained autoencoder) contents of arbitrary shapes at random locations, and 2) the corresponding masks.
It  imitates % formulates 
the noise pattern discrepancies between the real and synthesized contents in a more principled manner, without using any specific deep inpainting methods.
%In this way, our model can be trained with a large amount of inpainted images based on real images. 
% Therefore, our detection network remains effective when facing inpainted images manipulated by an unseen deep inpainting approaches, while existing detection algorithms suffer from a significant performance drop.
Second, we propose a Noise-Image Cross-fusion Network (NIX-Net) to effectively exploit the discriminative information contained in both the images and their noise residuals. After training on the universal dataset, our NIX-Net can reliably recognize the regions inpainted by different deep inpainting methods.
% (2)	To effectively exploit the discriminative information contained in both the image and its noise patterns, we propose a Noise-Image Cross-fusion Network (NIX-Net) for deep inpainting detection.
% In order to exchange the information across multi-scale feature representations, and yield rich feature representations for each scale with strengthened position sensitivity, we design a multi-scale cross fusion method to incorporate image and noise features.

In summary, our main contributions are:
\begin{itemize}
  \item We propose a novel framework for universal deep inpainting detection, which consists of 1) a new method of generating universal training data, and 2) a two-stream multi-scale Noise-Image Cross-fusion detection Network (NIX-Net). 

%   \item We present a novel way of generating universal training data %generation 
%   by utilizing the noise discrepancy between real image contents and deep inpainted contents.
  
%   \item We design a Noise-Image Cross-fusion Network for deep inpainting detection to better exploit the discriminative information contained in both the image and its noise residual.

  \item We empirically show, on multiple benchmark datasets, that our proposed approach can consistently outperform existing detection methods, especially when applied to detect unseen deep inpainting techniques.
%   in comparison with other detection approaches.

\item Our universal training dataset can also improve the generalizability of existing detection methods, making it an indispensable part of future detection methods.
\end{itemize}

\section{Related Work}

\subsection{Deep Image Inpainting}
Different from conventional image inpainting approaches, deep learning based image inpainting (or deep inpainting for short) trains inpainting networks on large-scale datasets and can generate more visually plausible details or fill large missing regions with new contents that never exist in the input image. 
By far, the generative adversarial networks (GAN) \cite{goodfellow2014generative} based inpainting methods are arguably the most powerful methods for deep image inpainting. These methods all employ a GAN-based training approach with two sub-networks: an inpainting network and a discriminative network. The former learns image semantics and fills the missing regions with predicted contents, whereas the latter distinguishes whether the image is real or inpainted. 
Phatak et al. \shortcite{pathak2016context} proposed the Context-Encoder (CE) for single image inpainting, which is known as the first GAN-based image inpainting technique. 
This technique was later improved by \cite{iizuka2017globally} using dilated convolution and global-local adversarial training. 
Yu et al. \shortcite{yu2018generative} proposed a two-stage inpainting network with a coarse-to-fine learning strategy.
% In particular, contextual attention is introduced to  borrowing pixels from distant locations within the same image to fill missing regions.
%Zeng et al. \shortcite{zeng2019learning} extends this idea by proposing a pyramid of contextual attention at multiple scales.
%Liu et al. \shortcite{liu2018image} propose the partial convolutional layer which masks the convolution operation and updates the mask in each layer, hence the prediction of the missing pixels only depends on the non-missing pixels in the original image.
A gated convolution network along with a learnable dynamic feature selection mechanism (for each channel and at each spatial location) was proposed in \cite{Yu_2019_ICCV} for image inpainting.
Li et al. \shortcite{Li_2020_CVPR} devised the Recurrent Feature Reasoning network  which recurrently enriches information for the hole region.
%Other deep inpainting approaches include the use of partial convolutional layer \cite{liu2018image} or multi-scale pyramid contextual attention  \cite{zeng2019learning}.
% Yu et al. proposed the gated convolution \shortcite{Yu_2019_ICCV} by devising a learnable dynamic feature selection mechanism .
Despite the diversity of existing deep inpainting methods, they all share a common characteristic: the noise patterns in generated contents are different from those in the real image contents. While this characteristic has been observed in understanding the artificial fingerprints of GANs~\cite{8695364}, it has not been exploited for deep inpainting detection. In this paper, we will leverage such a universal characteristic of generated contents to build universal deep inpainting detectors.  
% We hence aim to exploit this noise discrepancy in deep inpainted images to realize universal detection.

\subsection{Inpainting Forensics}
Deep inpainting detection falls into the general scope of image forensics, but quite different from the conventional image manipulation detection or deepfake detection.
Conventional image manipulation detection deals with traditional image forgery operations such as splicing \cite{huh2018fighting} and copy-move \cite{wu2018busternet}. 
Deepfake (or deep face swapping) is the other type of deep learning forgery techniques that swaps one person's face in a video to that of a different person, which often requires heavy post-processing including color transfer and boundary blending \cite{li2020face}. 
% As such, deepfake  Deepfake is mostly used for fake swap taking two 
Different from conventional image manipulation or deepfake, deep inpainting takes one image and a mask as inputs and generates new content for the mask region, based on information of the non-mask regions within the same image.
%Although the resulting contents may all appear realistic, different types of forgery techniques (eg. conventional image manipulation, deepfake or deep inpainting) generally require different detection techniques.
%Detectors developed for one type of forgery may not transfer to others. 
In this paper, we focus on deep image inpainting detection and the generalizability of the detector to unseen deep inpainting techniques (not to conventional image manipulation or deepfake).

Most of existing inpainting forensic methods are developed to detect traditional image inpainting techniques. For example, the detection of traditional diffusion-based inpainting based on local variance of image Laplacian \cite{li2017localization}, and the detection of traditional patch-based inpainting via patch similarities computed by zero-connectivity length \cite{wu2008detection}, two-stage suspicious region search \cite{chang2013forgery} or CNN-based encoder-decoder detection networks \cite{Zhu2018ADL}.
% Recent years have seen several approaches studying the forensics of conventional inpainting approaches. 
% Li et al. \shortcite{li2017localization} present a model to detect diffusion-based inpainting based on local variance of image Laplacian along the isophote direction. 
% To deal with patch-based inpainting, Wu et al. 
% \shortcite{wu2008detection} capitalize on the patch similarity computed by zero-connectivity length. 
% Bacchuwar et al. \cite{bacchuwar2013jump} devise a jump patch match approach to reduce the computational complexity. 
% Both \cite{wu2008detection} and \cite{bacchuwar2013jump} ask for the manual selection of suspicious regions and struggle with high false alarm rate. 
% Chang et al. \shortcite{chang2013forgery} present a two-stage searching algorithm to locate the suspicious regions and utilize multi-region relations to alleviate false alarms. 
% Besides, Liang et al. \shortcite{liang2015efficient} exploit central pixel mapping to speed up the search of suspicious regions.
% It also utilizes greatest zero-connectivity component labeling and fragment splicing detection to locate the tampered regions. 
% More recently, Zhu et al. \shortcite{Zhu2018ADL} devise a CNN-based encoder-decoder detection model to tackle patch-based inpainting for 256×256 images.
These methods are generally less effective on deep inpainting techniques that can synthesize extremely photo-realistic contents or new objects that never exist in the original image.
% However, compared with conventional inpainting approaches, deep generative inpainting approaches are capable of yielding more photo-realistic contents and even producing new objects in the missing regions, leading to different traces. 
% Hence, forensic approaches designed for conventional inpainting approaches are incompatible with the detection of deep generative inpainting.
Deep inpainting detection is a fairly new research topic.
The high-pass fully convolutional network by \cite{li2019localization} is the only known detection method for deep inpainting detection.
% Li et al.  made the first attempt on deep inpainting detection by using a high-pass fully convolutional network.
% \ma{Explain more about this method. 2-3 more sentences, explain how it works.}
However, training this detection network requires knowing the deep inpainting technique to be detected. This tends to limit its generalizability to unseen deep inpainting methods, as shown in our experiments. In this paper, we focus on developing universal deep inpainting detectors, which we believe is a crucial step towards more powerful and practical deep inpainting detectors.
% is trained on images tampered by specific deep inpainting approaches and thus fails to address the generalizability problem.
% In contrast, our proposed approach capitalizes on the common characteristic of deep inpainting approaches and thus enable universal detection.

%Considering that deep generative inpainting models are  trained with a generative adversarial process, researches on GAN generated image detection \cite{yu2019attributing,wang2020cnn} may have some relevance. 
%However, these approaches only classify whether an image is generated by a certain GAN model and cannot be applied to the task of deep inpainting detection.

\begin{figure}[t!]
\centering

\includegraphics[width=8.5cm]{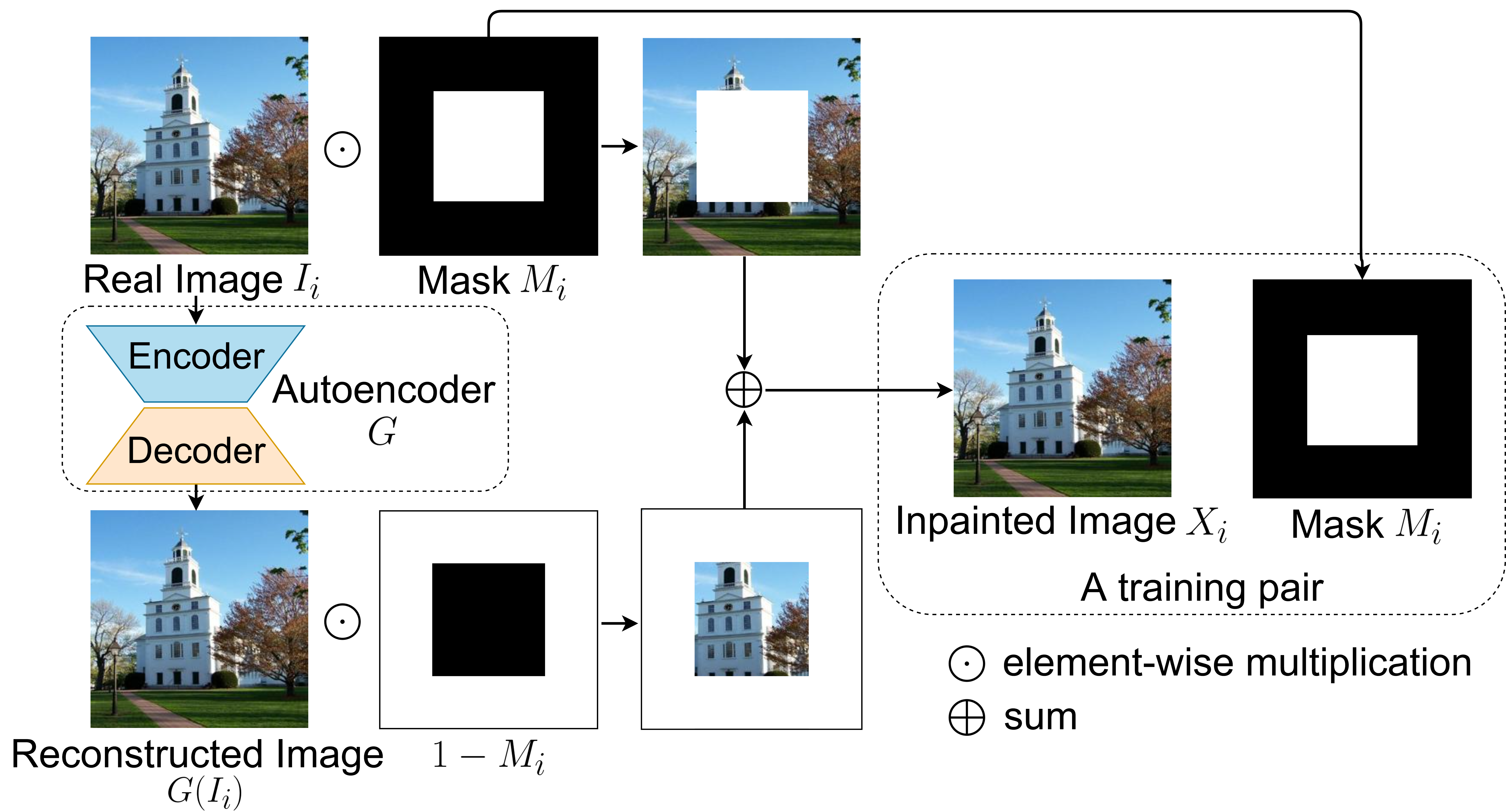}

\caption{The procedure of universal training dataset generation.}
\label{ijcai2021:combine} % I can do without the label too
\end{figure}

\begin{figure}[t!]
\centering

\includegraphics[width=6cm]{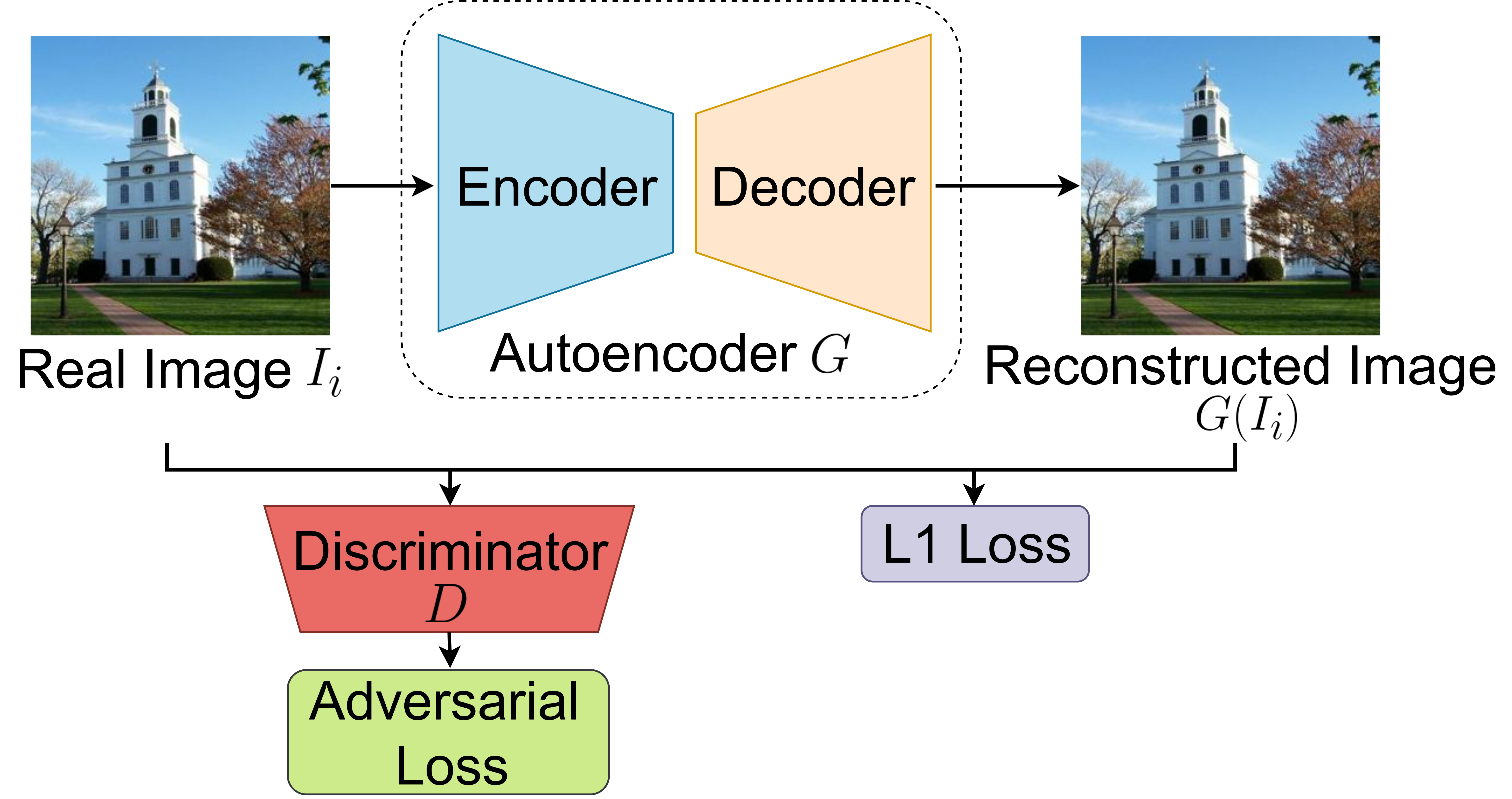}

\caption{The procedure of training the autoencoder $G$.}
\label{ijcai2021:EqualGAN} % I can do without the label too
\end{figure}

\section{Problem Formulation and Universal Training Dataset Generation}

\begin{figure*}[ht]
\centering
\includegraphics[width=0.88\linewidth]{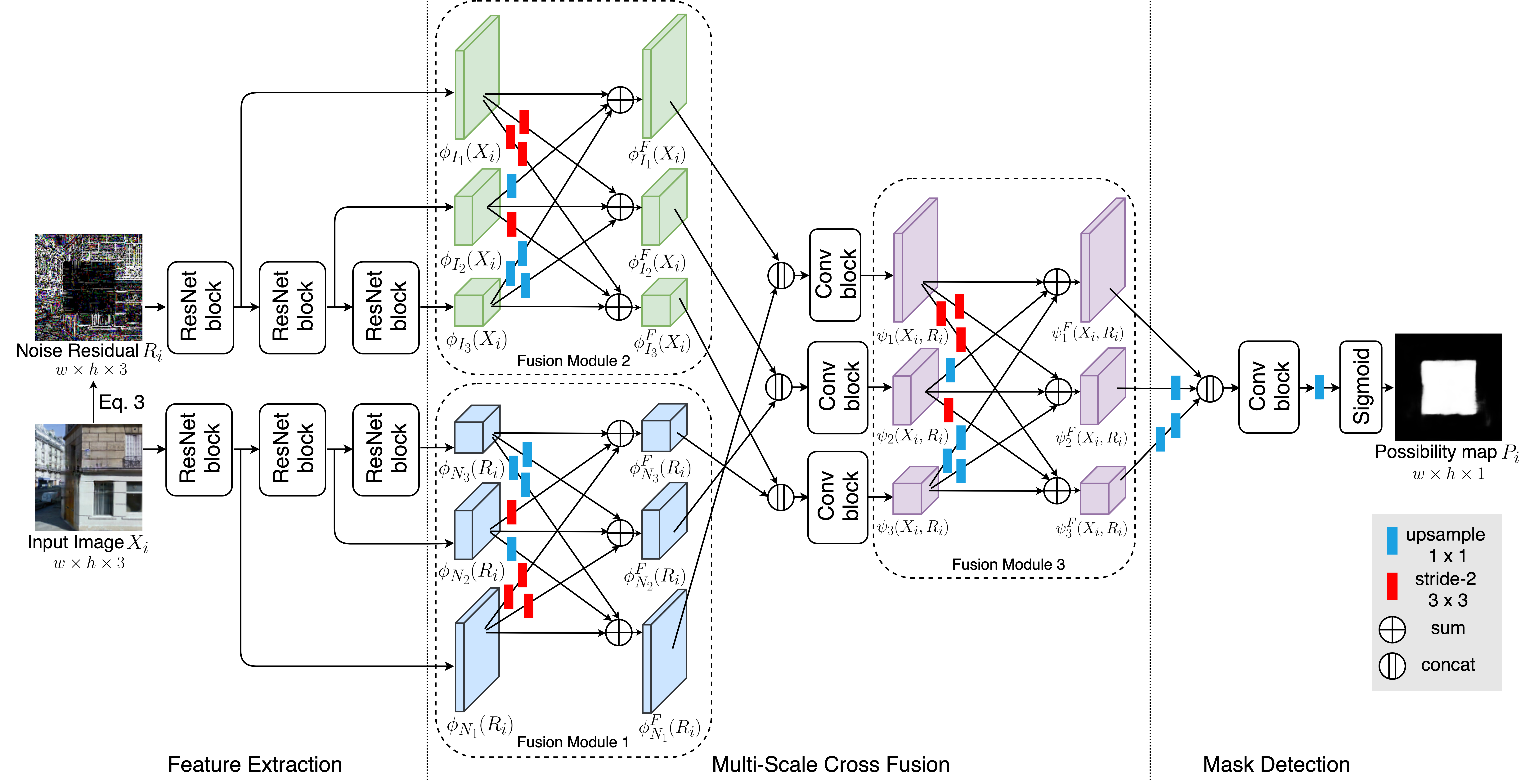}
\caption{Overview of the proposed Noise-Image Cross-Fusion Network (best viewed in color). Here ``upsample $1 \times 1$" and ``stride-2 $3 \times 3$" refer to bilinear upsampling followed by a $1 \times 1$ convolution, and $3 \times 3$ convolution with stride 2, respectively.}
\label{ijcai2021:detection}
\end{figure*}

\subsection{Problem Formulation}
Given an image $X_{i}$ inpainted by a certain deep inpainting method on regions defined by a binary mask $M_{i}$, deep inpainting detection aims to locate the inpainted regions $M_{i}$. A detection network can be trained to take the inpainted image $X_{i}$ as input and output the predicted mask $\hat{M}_{i}$.

% takes $X_{i}$ along with some other information given there as an inputs and outputs a binary mask $\hat{M}_{i}$.
% The problem is how to train a detection network to accurately predict the inpainted regions $M_{i}$.

 To train the detector, a straightforward approach is %to train the detection network on 
 to use inpainted images generated  by a deep inpainting method as the training data %to detect 
\cite{li2019localization}. 
However, this \emph{inpainting-method-aware} approach often generalizes poorly when applied to detect inpainted images generated by unseen deep inpainting methods. 
An empirically analysis can be found in Section~\ref{ijcai2021:sectionQuan}.
In contrast, we propose to generate a universal training dataset to capture the common characteristics shared by different deep inpainting methods, and train the detection model on this universal dataset. 
Such an \emph{inpainting-method-agnostic} approach can improve the generalization of the detection model to unseen deep inpainting methods. Next, we will introduce our proposed universal training dataset generation method inspired by the distinctive noise patterns exist in real versus generated contents.
% Existing detection models are mostly trained on images generated by a specific deep inpainting method. 
% Although decent detection accuracy is achieved on the test dataset generated by the same inpainitng approach, we observe significant degraded performance when applying the trained detection model to unseen inpainted images generated by other deep inpainting approaches, which is empirically verified in Section~\ref{ijcai2021:sectionQuan}.
% To improve the generalization ability of detection, we present a novel way %of 
% to generate universal training data to train the   detection network. 
%In other words, we do not require the prior knowledge of the deep inpainting approach that generates the testing data during training, which is more realistic in real-world scenarios.
%generation .

\subsection{Universal Training Dataset Generation}
\label{universal}
A universal training dataset should consider the common characteristics of different deep inpainting methods, rather than relying on the specific artifacts of one particular deep inpainting method. Motivated by our observation in Figure \ref{ijcai2021:noise}, here we propose to create the universal training dataset by simulating ``inpainted" images from autoencoder reconstructions, instead of using any existing deep inpainting methods. The complete generation procedure is illustrated in Figure~\ref{ijcai2021:combine}.
% In order to ensure high generalization ability of detection, the training data is supposed to embody the common characteristic of most deep inpainting approaches, rather than rely on the synthesized artifacts of specific ones.
% We are motivated from a decisive observation: when an image is manipulated by any deep inpainting approach, there exist intrinsic discrepancies on noise patterns between real image contents and those generated by deep inpaiting.   
% Specifically, % this image contains two types of underlying noise patterns: 
%(1) 
% the undamaged regions, due to the acquisition process, contain noise marks which are introduced from hardware (e.g., camera) and present similarly throughout the image \cite{Farid2009}.
%(2)
% In contrast, the inpainted regions 
% are left with unique noise patterns by GAN-based inpainting algorithms \cite{8695364,yu2019attributing}.
% As a consequence, we propose to generate a universal training dataset for deep inpainting detection, by exploiting the inconsistency of the underlying noise patterns between the undamaged regions and the inpainted regions.
% To be specific, we imitate the noise inconsistency %above and 
% to generate %simulate 
% inpainted images as training samples for deep inpainting detection.
% The overall procedure is illustrated in Figure~\ref{ijcai2021:combine}.

For a set of real images $I \coloneqq \{I_{1},I_{2},...,I_{n}\}$, the generation process generates a set of simulated (as it does not use any real inpainting methods) inpainted images $X \coloneqq \{X_{1},X_{2},...,X_{n}\}$ using a pre-trained autoencoder with a set of random binary masks $M \coloneqq \{M_{1},M_{2},...,M_{n}\}$. 
%The generated images and the random masks together form the universal training dataset.  
Specifically, given a real image $I_{i}$, we first obtain its reconstructed version $G(I_{i})$ from an autoencoder $G$. 
We train the autoencoder following a typical GAN~\cite{goodfellow2014generative} approach using the autoencoder as the GAN generator and an additional classification network as the GAN discriminator. The overall structure is illustrated in Figure \ref{ijcai2021:EqualGAN}. The autoencoder is trained to have small reconstruction error, and at the same time, the reconstructed images should be as realistic as the real images according to the discriminator.
% generate  reconstruction $G(I_{i})$ that is as similar as possible to the input real image $I_{i}$, according to the discriminator. 
The overall training loss of this autoencoder is:
\begin{equation}
%\begin{split}
    \mathcal{L} = \sum_{i=1}^{n}\underbrace{\log (D(I_{i}) + \log (1-D(G(I_{i})))}_\text{$\mathcal{L}_{\text{adv}}$} + \lambda \underbrace{\Vert G(I_{i}) - I_{i} \Vert_{2}}_\text{$\mathcal{L}_{\text{rec}}$}
%\end{split}
\end{equation}
where, $\mathcal{L}_{\text{adv}}$ and $\mathcal{L}_{\text{rec}}$ are the adversarial and reconstruction loss respectively, and $\lambda=0.1$ is a trade-off parameter.

After training, the autoencoder $G$ is applied to reconstruct each real image in $I$.
With the reconstructed images, our next step is to simulate the inpainting process.  Specifically, we simulate an ``inpainted'' image $X_{i}$ by combining $I_{i}$ and its reconstruction $G(I_{i})$ according to a random mask $M_i$:
\begin{equation}
    X_{i} =  M_{i} \odot I_{i} + (1-M_{i}) \odot G(I_{i}),
\end{equation}
where $\odot$ is the element-wise multiplication and $M_{i}$ is a binary mask with 0 elements indicating the inpainting region (white region) and 1 elements indicating the non-inpainting region (black region).
%We generate random masks with arbitrary shapes at random locations of the input space (more detail can be found in the experiment section).
In $X_{i}$, the ``inpainted" region carries over the noise patterns of the synthesized contents from the autoencoder, while the rest of the regions preserve the noise patterns from the real contents.
Following this procedure, we can obtain a set of images with synthesized regions to create the universal training dataset: $UT = \{X, M\}$, with $X$ being the simulated inpainted images and $M$ being the inpainting masks.
% In this way, the generation of $X_{i}$ successfully imitates the noise discrepancy of deep inpainted images, without needing to access any specific deep inpainting approaches.
% Finally, a training pair for deep inpainting detection is obtained, which includes the inpainted image $X_{i}$ and its corresponding inpainting mask $M_{i}$.
% Therefore, we are capable of producing a large number of training data with only real images and without any deep inpainting approaches.
% Specifically, let the generated Universal Training dataset for deep inpainting detection be $UT = \{X, M\}$ where $X$ represents the inpainted image and $M$ denotes the corresponding inpainting mask.

Our universal dataset $UT$ distinguishes itself from existing inpainting-method-aware datasets as a general formulation of real versus generated contents.
% In other words, it does not depend on any inpainting method. 
However, having this dataset is not enough to train accurate deep inpainting detectors.
It requires an effective learning framework to exploit the discriminative information contained in the dataset, especially the noise patterns.
Next we will introduce our proposed detection network that serves this purpose.

\section{Noise-Image Cross-fusion Network (NIX-Net)}
%Figure~\ref{ijcai2021:detection} illustrates the proposed Noise-Image Cross-fusion Network (NIX-Net) for deep inpainting detection. 
As shown in Figure~\ref{ijcai2021:noise} (c), the  noise residual patterns   between  inpainted and non-inpainted regions in an inpainted image are distinct.
Unlike previous works that ignore this important cue  for mask detection,
we propose NIX-Net which leverages both the inpainted image $X_{i}$ and its noise residual $R_i$ %between $X_{i}$ and its denoised version $d(X_{i})$ by an additional denoiser $d(\cdot)$ 
%s inputs
to enhance detection performance.
%Note that NIX-Net is different from standard detection network which only takes an inpainted image as input and output the predicted mask. 
% The noise residual is defined as the pixel-wise difference between $X_{i}$ and its denoised version $f(X_{i})$ by a filter. 
%The need of an additional denoiser $f(\cdot)$ rather than directly using ground truth $I_i$ or the autoencoder pre-trained for the dataset generation is that, at test time, we do not have the ground truth $I_i$ nor a denoising autoencoder pre-trained on unseen datasets.
As shown in Figure~\ref{ijcai2021:detection}, the proposed NIX-Net consists of three components: 1) feature extraction, 2) multi-scale cross fusion, and 3) mask detection.  %The detailed designs of the three components are as follows.
% Given an inpainted image $X_{i}$, its underlying noise pattern stands for a kind of  disturbance, which is irrelevant to the image semantics.
% However, the above observation  indicates that the noise pattern can provide strong cues to unveil traces of deep inpainting.
% Therefore, we propose a
% %the 
% Noise-Image Cross-fusion Network (NIX-Net) for deep inpainting detection, which aims to effectively integrate the discriminative information contained in both the image and its noise pattern.
% As shown in Figure~\ref{ijcai2021:detection}, NIX-Net consists of three components, \textit{i.e.}, feature extraction, multi-scale cross fusion, and mask detection.

\paragraph{Feature extraction.} Given an inpainted image $X_i$, we define its noise residual as following:
\begin{equation}
    R_{i} = X_{i} - d(X_{i}),
\end{equation}
Inspired by recent studies on image forensics using SRM features \cite{zhou2018learning}, we choose the SRM filter as our denoising filter $d(\cdot)$. 
The feature extraction component includes two parallel feature extraction streams to learn multi-scale feature maps of the input image and its noise residual, respectively. %As shown in Figure~\ref{ijcai2021:detection}, 
Each feature extraction module consists of three ResNet blocks \cite{he2016identity}, resulting in 
three feature maps of $X_{i}$, namely, $\phi_{I_{1}}(X_{i})$, $\phi_{I_{2}}(X_{i})$ and $\phi_{I_{3}}(X_{i})$. The numbers of channels in the feature maps  are 128, 256 and  512, respectively.
Likewise, the three feature maps of $R_{i}$ are $\phi_{N_{1}}(R_{i})$, $\phi_{N_{2}}(R_{i})$ and $\phi_{N_{3}}(R_{i})$.
Since the last convolution layer of each ResNet block has a stride of 2 to reduce the spatial scale \cite{he2016deep}, %we can obtain three feature maps of $X_{i}$, namely, $\phi_{I_{1}}(X_{i})$, $\phi_{I_{2}}(X_{i})$ and $\phi_{I_{3}}(X_{i})$.
the spatial scales of the feature maps are $1/2, 1/4$ and $1/8$ of the input spatial size. %, and the numbers of feature map channels are 128, 256, 512.
%Likewise, the three feature maps of $R_{i}$ are $\phi_{N_{1}}(R_{i})$, $\phi_{N_{2}}(R_{i})$ and $\phi_{N_{3}}(R_{i})$.

% We first extract the noise pattern of the input image, namely, the noise residual.
% Specifically, we obtain the denoised image $f(X_{i})$ through an appropriate denoising filter $f()$, then the noise residual can be extracted by subtracting the denoised image from the original one:

% \begin{equation}
%     R_{i} = X_{i} - f(X_{i}),
% \end{equation}

\paragraph{Multi-scale cross fusion.}
The multi-scale cross fusion component aims to effectively incorporate noise features and image features in a multi-scale manner.
Specifically, both the noise stream and the image stream are followed by a fusion module. 
The fusion module takes the three-scale feature maps as input, and  outputs the crossly fused three-scale feature maps.
Each fused feature map is the sum of the three transformed (upsampled, downsampled or unchanged) input feature maps.
The outputs of fusion module 1 in the image stream are $\phi_{I_{1}}^{F}(X_{i})$, $\phi_{I_{2}}^{F}(X_{i})$ and $\phi_{I_{3}}^{F}(X_{i})$, while the outputs of fusion module 2 in the noise stream are $\phi_{N_{1}}^{F}(R_{i})$, $\phi_{N_{2}}^{F}(R_{i})$ and $\phi_{N_{3}}^{F}(R_{i})$.
The purpose of this fusion module is to exchange the information across multi-scale feature representations, and produce richer feature representations with strengthened position sensitivity ~\cite{Sun_2019_CVPR}.
For each scale, we concatenate the feature maps of the image stream and the noise stream along the channel dimension, followed by a Conv block.
%After feeding the concatenated feature maps to a Conv block, we obtain
The outputs are denoted as $\psi_{1}(X_{i},R_{i})$, $\psi_{2}(X_{i},R_{i})$ and $\psi_{3}(X_{i},R_{i})$.
The Conv block consists of two $3 \times 3$ convolutions, each of which is followed by a batch normalization and a ReLU Layer.
Finally, fusion module 3 further consolidates the connection between the image and noise features over different scales, and output three fused feature maps $\psi_{1}^{F}(X_{i},R_{i})$, $\psi_{2}^{F}(X_{i},R_{i})$ and $\psi_{3}^{F}(X_{i},R_{i})$.

\paragraph{Mask detection.}
The mask detection module first upsamples the two lower-resolution feature maps $\psi_{2}^{F}(X_{i},R_{i})$ and $\psi_{3}^{F}(X_{i},R_{i})$ using bilinear upsampling so that they have the same resolution as $\psi_{1}^{F}(X_{i},R_{i})$.
%to the high resolution 
%through bilinear upsampling with the same number of channels,
The three feature maps are then concatenated along the channel dimension, followed by a Conv block and an upsampling layer which outputs a $w \times h \times 1$ feature map.
%It then concatenates the three feature maps along the channel dimension.
The feature map is then fed into a Sigmoid layer for classification, rendering the possibility map $P_{i}$ with pixel-wise predictions.
Finally, the detected mask $\hat{M}_{i}$ can be obtained by binarizing $P_{i}$ according to a threshold value.
In this paper, we set the threshold value as 0.5.

\paragraph{Network training.} We train the entire network end-to-end using the focal loss \cite{lin2017focal} on the universal training dataset $UT$. Note that the network can also be trained on any other inpainting detection datasets including the inpainting-method-aware datasets. The use of the focal loss is to mitigate the effect of class imbalance (the inpainted regions are often small compared to the entire image). The focal loss $\mathcal{L}_{f}$ is defined as following:
\begin{equation}
\resizebox{.9\hsize}{!}{$
\mathcal{L}_{f} = \sum_{i}^{n}(-M_{i} (1-\hat{M}_{i})^{\gamma }\log \hat{M}_{i} - (1-M_{i})\hat{M}_{i}^{\gamma }\log (1- \hat{M}_{i}))
$
}
\end{equation}
where $\gamma$ is the focusing parameter and is set to 2.
% For the loss function of our proposed detection network, we employ  which is modified based on the standard binary cross entropy loss and can mitigate the effect of class imbalance. 
% It attaches a modulating factor to the cross entropy term, so as to decrease the weights for the dominant and well-classified negative samples in the total loss. 

\section{Experiments}
In this section, we first introduce the experimental settings, then evaluate the performance of our proposed approach via extensive experiments and ablation studies.

\paragraph{Inpainting methods and datasets.}
We use three different deep inpainting techniques including \textbf{GL} \cite{iizuka2017globally}, \textbf{CA} \cite{yu2018generative} and \textbf{GC} \cite{Yu_2019_ICCV} to generate inpainted images on two datasets Places2~\cite{zhou2017places} and CelebA~\cite{liu2015deep}.  
For each of the two datasets, we randomly select (without replacement) 50K, 10K and 10K images to create the training, validation and testing subsets respectively, following either our universal data generation or using one of the above three inpainting techniques (GL, CA and GC).
We train the detection models on the training subset and test their performance on the test subset.
%The universal training datasets are also generated for both Places2 and CelebA (from their training subsets) using our proposed approach in Section \ref{universal}.
% Besides, we prepare extra training datasets by exploiting Places2 and CelebA  datasets with our proposed universal training data generation approach in Section~\ref{universal}, which are denoted as \textbf{UT}.

\begin{figure*}[t!]
%\vspace{-0.05 in}
    \centering
    \setlength{\tabcolsep}{0.05em}
    {
        \begin{tabular}{cccccc @{\hspace{0.15cm}} cccccc}

\includegraphics[width=1.44cm]{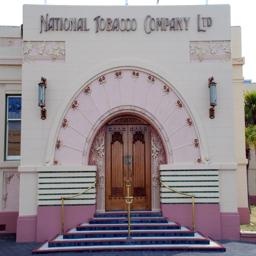}&
\includegraphics[width=1.44cm]{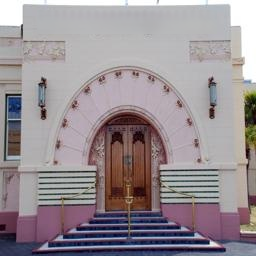}&
\includegraphics[width=1.44cm]{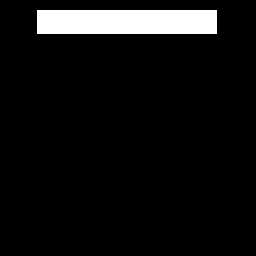}&
\includegraphics[width=1.44cm]{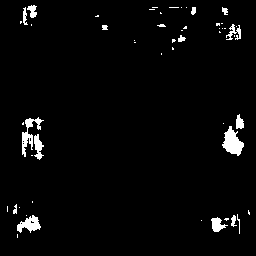}&
\includegraphics[width=1.44cm]{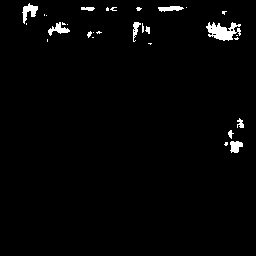}&
\includegraphics[width=1.44cm]{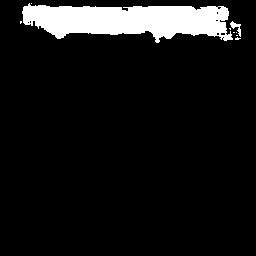}&

\includegraphics[width=1.44cm]{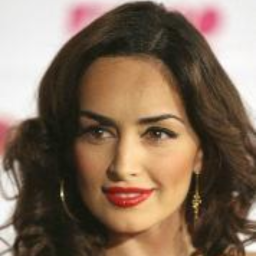}&
\includegraphics[width=1.44cm]{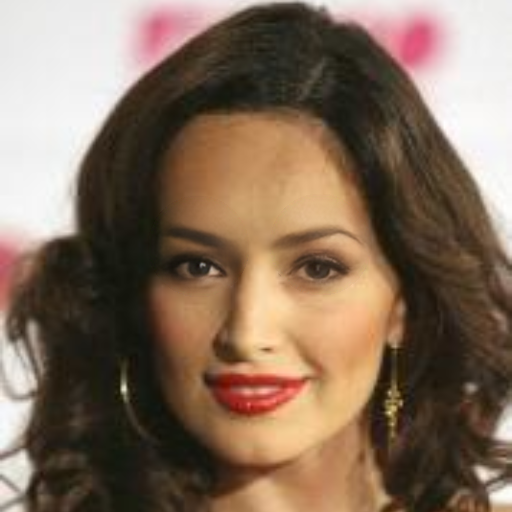}&
\includegraphics[width=1.44cm]{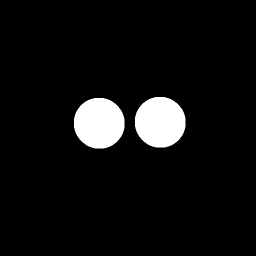}&
\includegraphics[width=1.44cm]{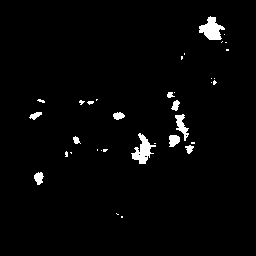}&
\includegraphics[width=1.44cm]{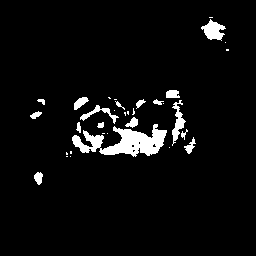}&
\includegraphics[width=1.44cm]{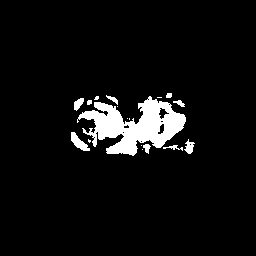}
\\
\addlinespace[-0.23em]

\includegraphics[width=1.44cm]{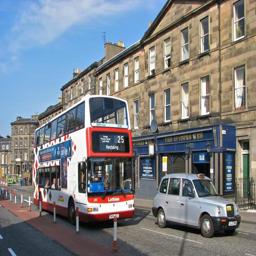}&
\includegraphics[width=1.44cm]{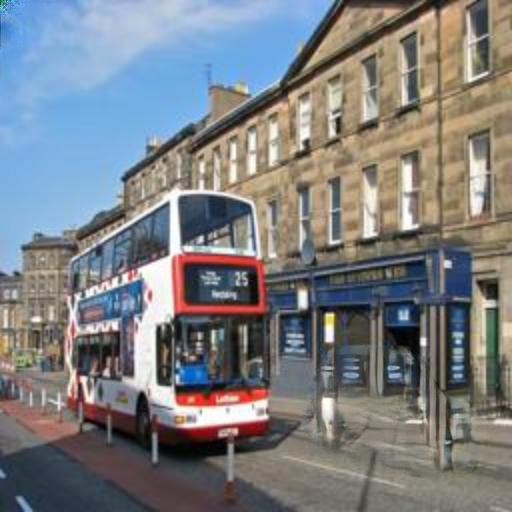}&
\includegraphics[width=1.44cm]{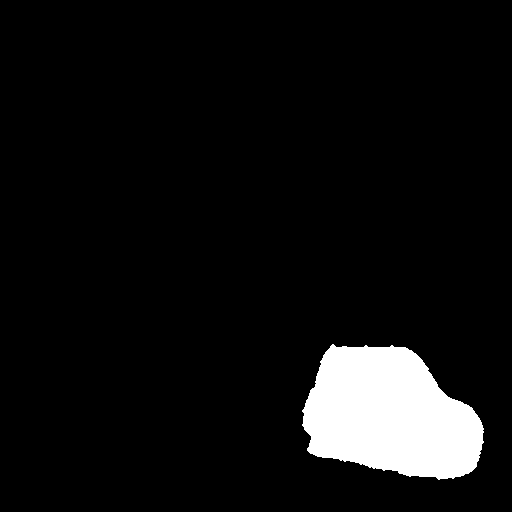}&
\includegraphics[width=1.44cm]{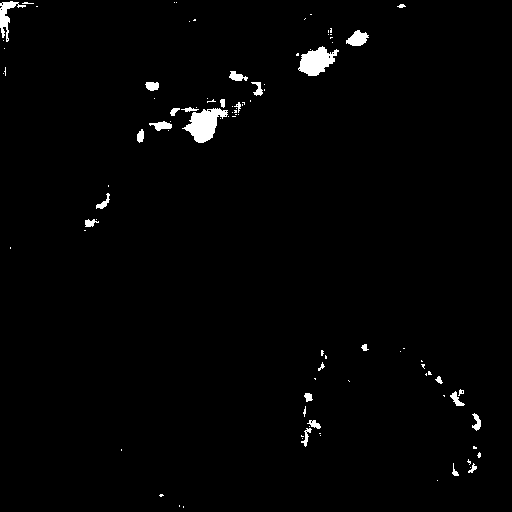}&
\includegraphics[width=1.44cm]{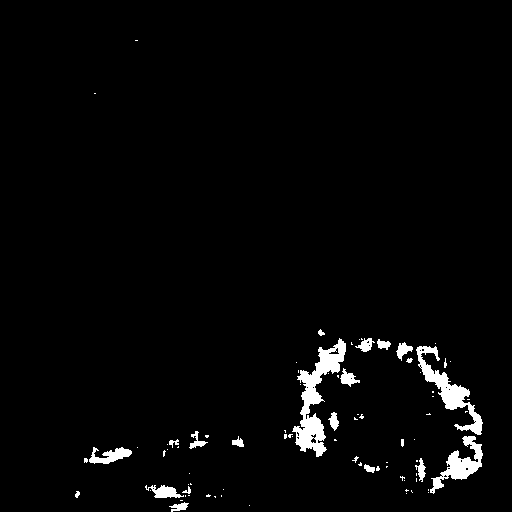}&
\includegraphics[width=1.44cm]{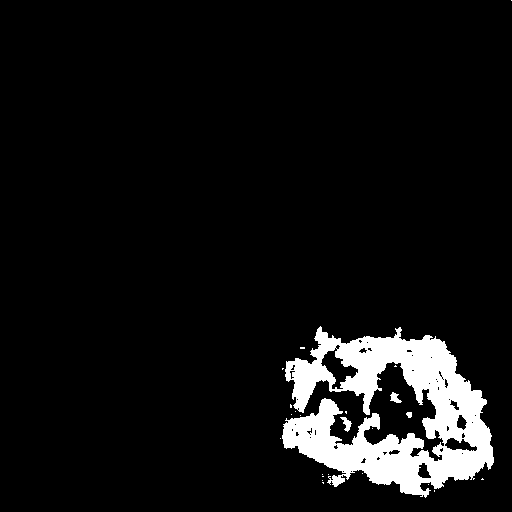}& 

\includegraphics[width=1.44cm]{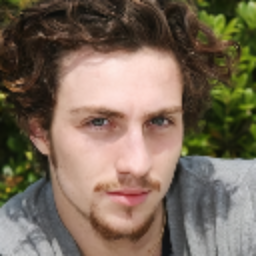}&
\includegraphics[width=1.44cm]{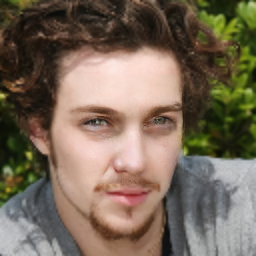}&
\includegraphics[width=1.44cm]{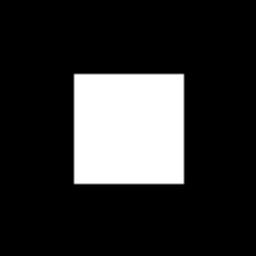}&
\includegraphics[width=1.44cm]{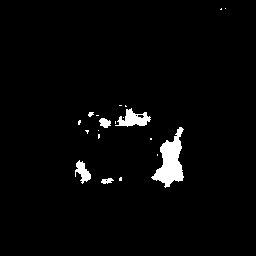}&
\includegraphics[width=1.44cm]{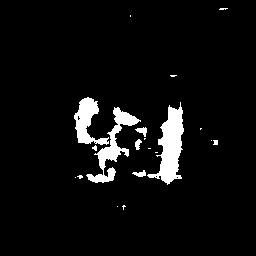}&
\includegraphics[width=1.44cm]{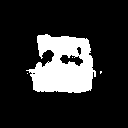}
\\
\addlinespace[-0.24em]

\includegraphics[width=1.44cm]{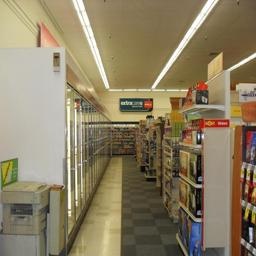}&
\includegraphics[width=1.44cm]{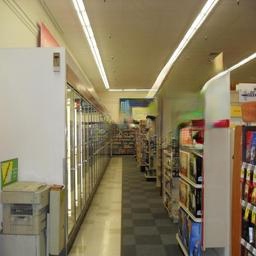}&
\includegraphics[width=1.44cm]{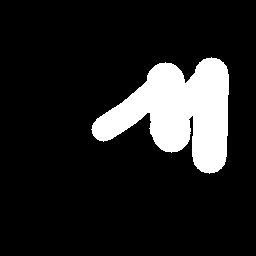}&
\includegraphics[width=1.44cm]{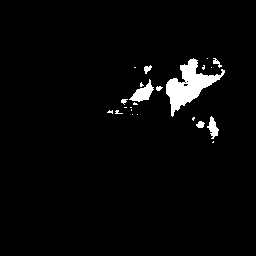}&
\includegraphics[width=1.44cm]{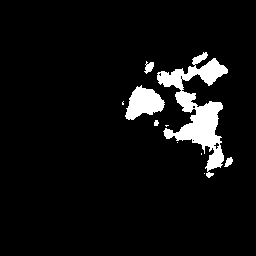}&
\includegraphics[width=1.44cm]{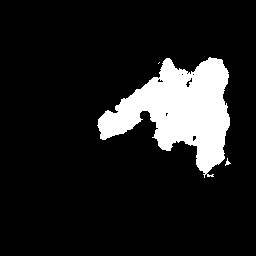}&

\includegraphics[width=1.44cm]{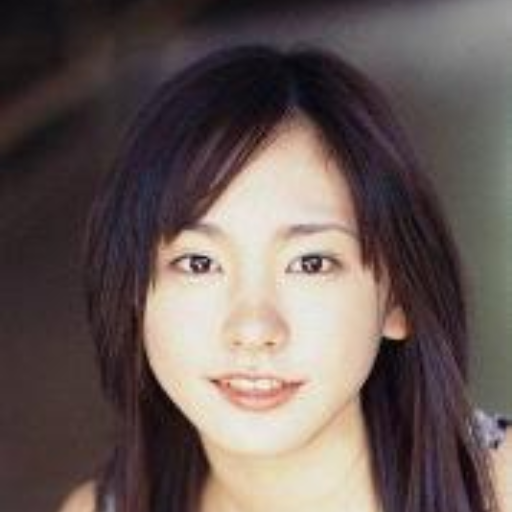}&
\includegraphics[width=1.44cm]{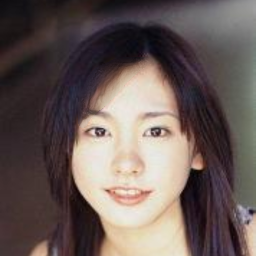}&
\includegraphics[width=1.44cm]{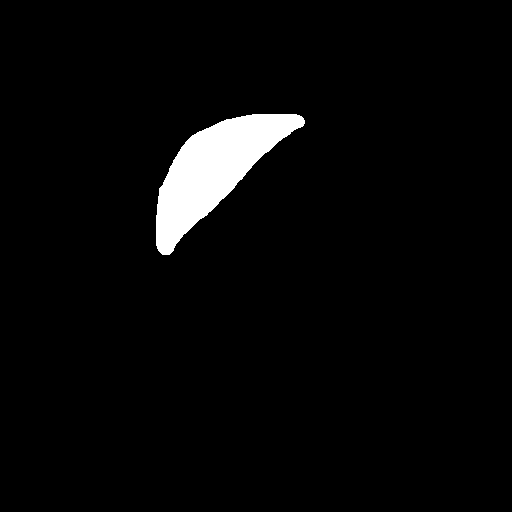}&
\includegraphics[width=1.44cm]{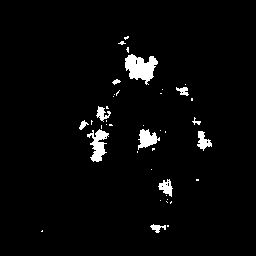}&
\includegraphics[width=1.44cm]{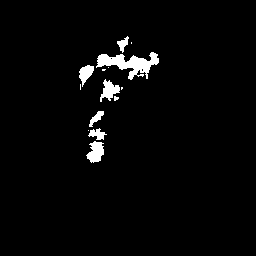}&
\includegraphics[width=1.44cm]{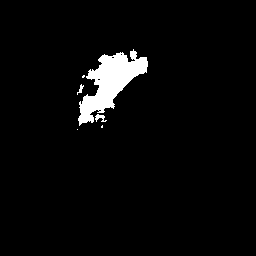}
\\
\addlinespace[-0.24em]

\includegraphics[width=1.44cm]{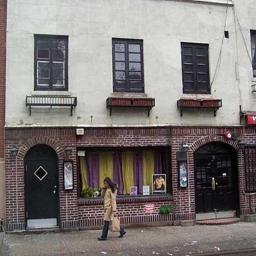}&
\includegraphics[width=1.44cm]{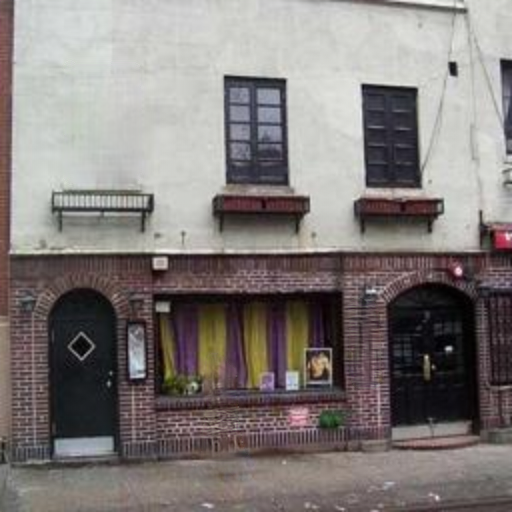}&
\includegraphics[width=1.44cm]{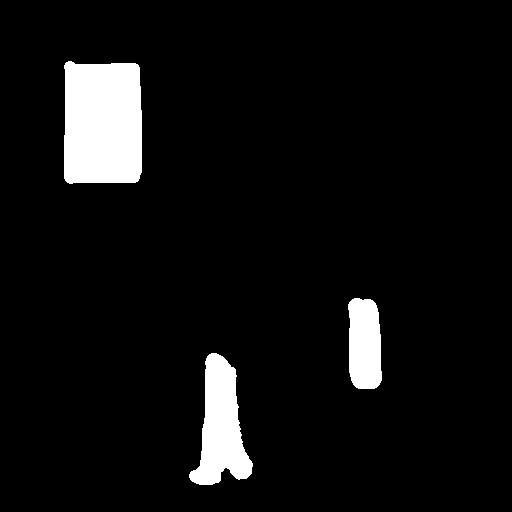}&
\includegraphics[width=1.44cm]{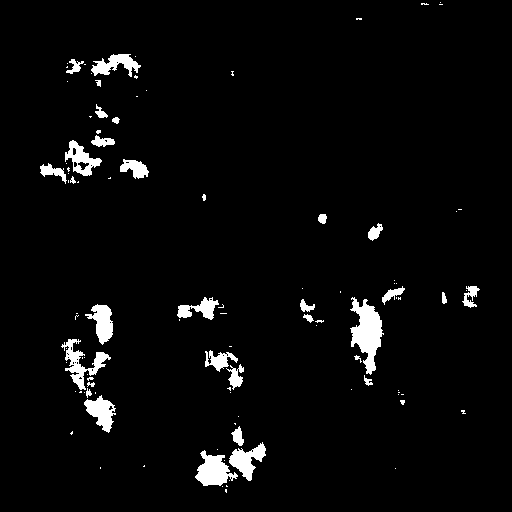}&
\includegraphics[width=1.44cm]{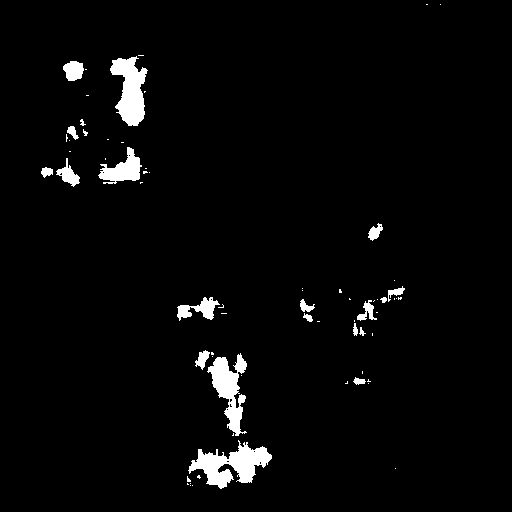}&
\includegraphics[width=1.44cm]{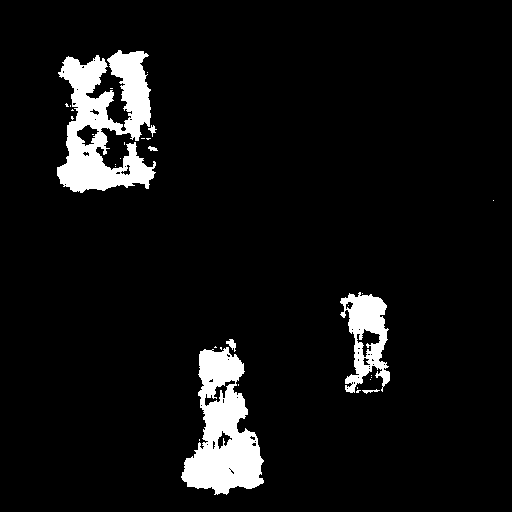}&

\includegraphics[width=1.44cm]{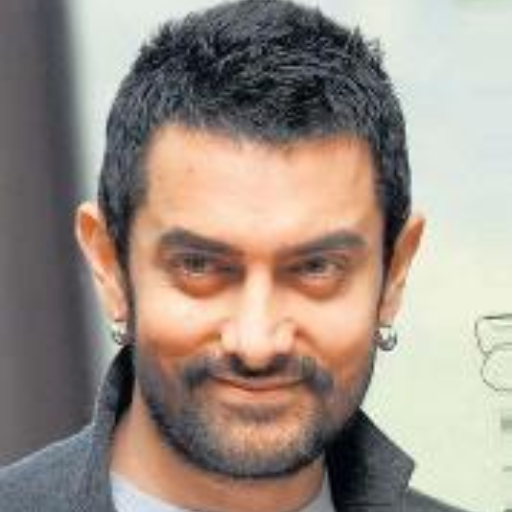}&
\includegraphics[width=1.44cm]{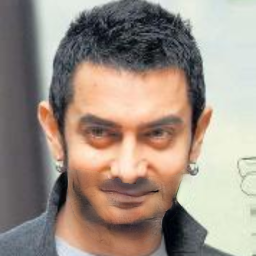}&
\includegraphics[width=1.44cm]{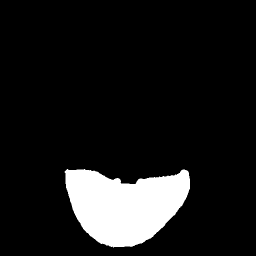}&
\includegraphics[width=1.44cm]{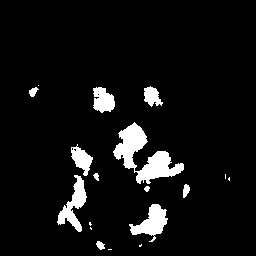}&
\includegraphics[width=1.44cm]{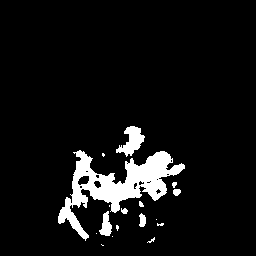}&
\includegraphics[width=1.44cm]{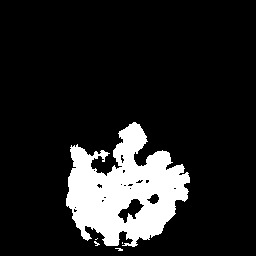}
\\
\addlinespace[-0.23em]

\includegraphics[width=1.44cm]{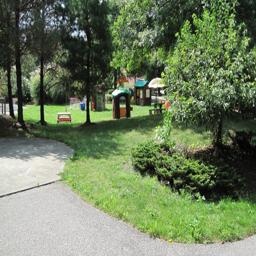}&
\includegraphics[width=1.44cm]{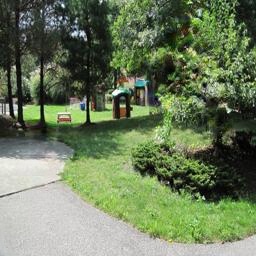}&
\includegraphics[width=1.44cm]{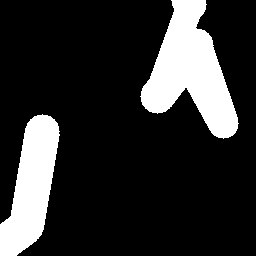}&
\includegraphics[width=1.44cm]{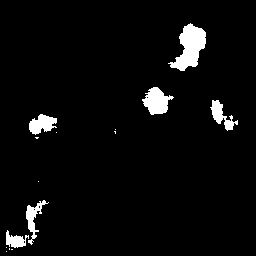}&
\includegraphics[width=1.44cm]{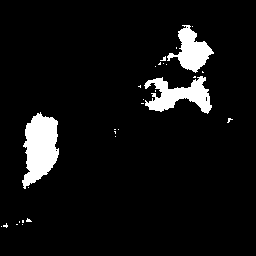}&
\includegraphics[width=1.44cm]{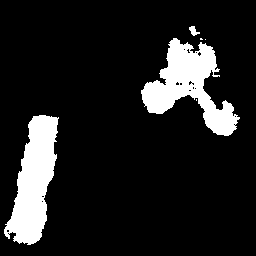}&

\includegraphics[width=1.44cm]{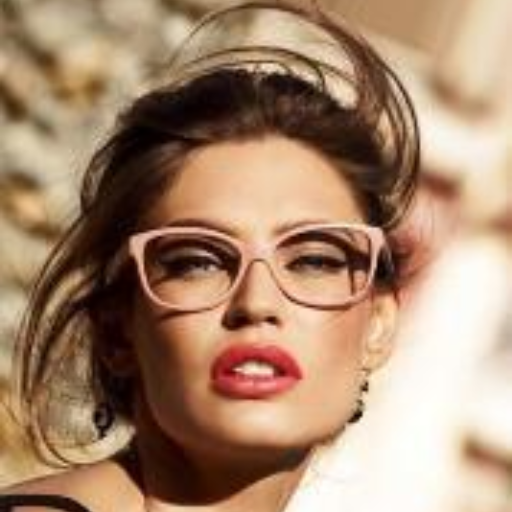}&
\includegraphics[width=1.44cm]{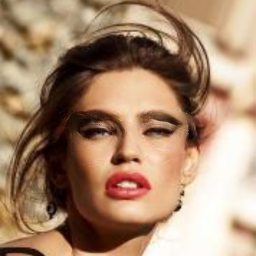}&
\includegraphics[width=1.44cm]{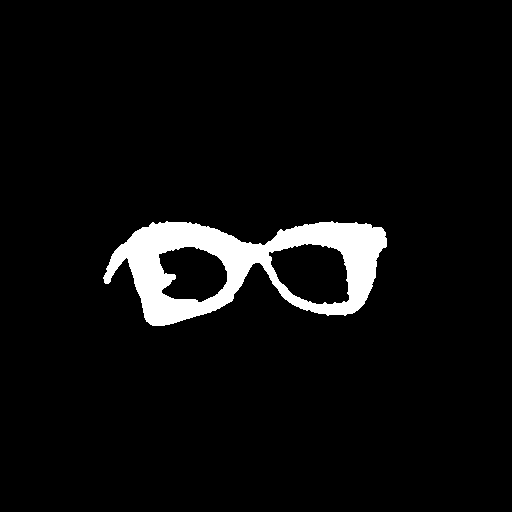}&
\includegraphics[width=1.44cm]{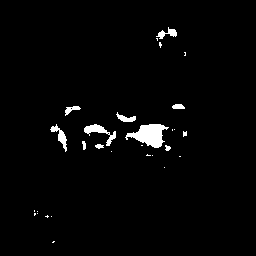}&
\includegraphics[width=1.44cm]{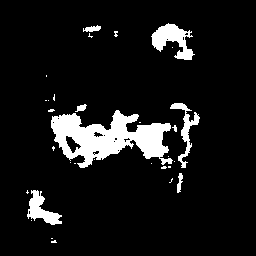}&
\includegraphics[width=1.44cm]{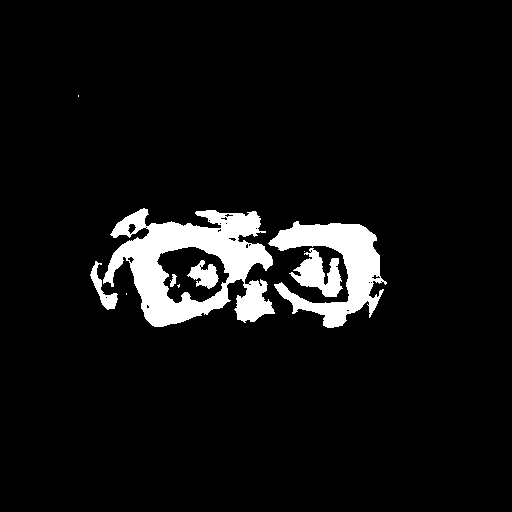}
\\

\scriptsize Original & \scriptsize Inpainted &\scriptsize  Mask GT &\scriptsize  LDICN  &\scriptsize  ManTra-Net &\scriptsize  Ours & \scriptsize Original & \scriptsize Inpainted &\scriptsize  Mask GT &\scriptsize  LDICN  &\scriptsize  ManTra-Net &\scriptsize  Ours\\

\end{tabular}

}
    
\caption{Qualitative comparisons on Places2 and CelebA. The original images are inpainted by CA. Mask GT refers to the ground truth of inpainting mask. LDICN and ManTra-Net are only trained on data generated by GL. Our model is only trained with UT data.}
\label{ijcai2021:QualitativeCompare1}
% \vspace{-0.4cm}
\end{figure*}

\paragraph{Mask generation.}
To simulate more diverse and complex real-world scenarios, we utilize the irregular mask setting in \cite{Yu_2019_ICCV} with arbitrary shapes and random locations for both training and testing.
Besides, object-shape masks are also adopted for visual comparison, as shown in Figure~\ref{ijcai2021:QualitativeCompare1}.

\paragraph{Baseline models.}
We consider two baseline models:
1) \textbf{LDICN} \cite{li2019localization}, a fully convolutional network designed for deep inpainting detection; and
2) \textbf{ManTra-Net} \cite{Wu_2019_CVPR}, a state-of-the-art detection model for traditional image forgery such as splicing. 
% We retrain it for deep inpainting detection.

\paragraph{Performance metric.}
We use the Intersection over Union (IoU) as the performance metric, and report the mean IoU (mIoU) over the entire test subset of inpainted images.

\paragraph{Training setting.}
We train the networks using the Adam optimizer with initial learning rate $1\times10^{-4}$.
An early stopping strategy is also adopted based on the mIoU on the validation dataset: the model with the highest validation mIoU is saved as the final model. 
All of our experiments were run with a Nvidia Tesla V100 GPU.

\begin{table}[!htb]
\large
\renewcommand\arraystretch{1.05}
\begin{center}
\scalebox{0.71}
{
\begin{tabular}{l|cc|ccc|ccc}
      			
      	\toprule
      	 \multirow{3}{*}{\textbf{Model}}  &
      	 \multicolumn{2}{c|}{\textbf{Training}} & \multicolumn{6}{c}{\textbf{Test mIoU}} \\ \cline{4-9}
      	 & \multicolumn{2}{c|}{\textbf{Data}} & \multicolumn{3}{c|}{Places2} & \multicolumn{3}{c}{CelebA} \\ \cline{2-9}
     			& GL  & UT & GL & CA & GC & GL & CA & GC \\ \hline 
		LDICN   & \checkmark &   & 83.47 & 66.70 &  56.24 & 87.27 & 67.61 & 64.16\\	 
		ManTra-Net  & \checkmark &   & 88.76 & 70.18 & 64.60 & 92.53 & 76.22 & 70.98 \\ 
      		\textbf{NIX-Net} & \checkmark &   & \textbf{91.82} & \textbf{80.55} & \textbf{77.63}  & \textbf{93.37} & \textbf{84.48} & \textbf{81.24} \\ \hline
      		\textbf{NIX-Net} & \checkmark & \checkmark  & \textbf{92.14} & \textbf{86.09} & \textbf{81.98} & \textbf{93.71} & \textbf{89.63} & \textbf{87.95}  \\ 
      		
      		\hline \hline
      		
      		& CA  & UT & GL & CA & GC & GL & CA & GC\\ \hline
      		LDICN   & \checkmark &   & 69.53 & 82.48 & 57.73 & 75.85 & 87.04 & 68.49 \\	 
		ManTra-Net  & \checkmark &   & 76.22 & 86.08 & 69.61 & 81.21 & 89.40 & 77.39 \\ 
      		\textbf{NIX-Net} & \checkmark &   & \textbf{83.57} & \textbf{88.75} & \textbf{76.49}  & \textbf{87.93} & \textbf{92.30} & \textbf{83.77} \\ \hline
      		\textbf{NIX-Net} & \checkmark & \checkmark  & \textbf{90.50} & \textbf{89.16} & \textbf{83.80} & \textbf{92.49} & \textbf{92.74} & \textbf{88.36} \\ 
      		
      		\hline \hline
      		
      		& GC  & UT & GL & CA & GC & GL & CA & GC\\ \hline
      		LDICN   & \checkmark &   & 70.55 & 68.16 & 84.24 & 77.62 & 73.81 & 87.29 \\	 
		ManTra-Net  & \checkmark &   & 80.85 & 74.69 & 84.90 & 83.31 & 81.25 & 88.46 \\ 
      		\textbf{NIX-Net} & \checkmark &   & \textbf{84.77} & \textbf{81.03} & \textbf{85.38} & \textbf{90.57} & \textbf{86.44} & \textbf{88.97} \\ \hline
      		\textbf{NIX-Net} & \checkmark & \checkmark  & \textbf{91.48} & \textbf{87.25} & \textbf{85.61} & \textbf{93.11} & \textbf{91.82} & \textbf{90.34} \\ 
      		
      		\hline \hline
      		
      		&    & UT & GL & CA & GC & GL & CA & GC\\ \hline
      		LDICN   & & \checkmark    & 82.95 & 80.79 & 78.29 & 85.52 & 82.98 & 81.43  \\	 
		ManTra-Net  & & \checkmark   & 88.42 & 83.15 & 80.52 & 89.71 & 86.64 & 85.38 \\ 
      		\textbf{NIX-Net} & & \checkmark     & \textbf{91.33} & \textbf{88.46} & \textbf{84.71} & \textbf{93.06} & \textbf{91.59} & \textbf{88.20} \\ 
      		\bottomrule

\end{tabular}

}
\end{center}
\caption{Quantitative Comparison on Places2 and CelebA datasets.}
\label{ijcai2021:tab0}
\end{table}

\subsection{Quantitative Performance Evaluation}
\label{ijcai2021:sectionQuan}
We have 3 detection networks (LDICN, ManTra-Net and our NIX-Net) and 2 types of training data including 1) 3 \emph{inpainting-method-aware} training datasets created using 3 inpainting methods (GL, CA and GC), and 2) our \emph{universal} (UT) training dataset. Here, we first train the detection networks on GL/CA/GC then test the performance on the test sets of all three datasets. 
Besides, we run our NIX-Net on a hybrid dataset that combines UT with one out of the 3 \emph{inpainting-method-aware} datasets. 
We also train the detection networks on UT only to test the importance of universal training data to generalizability. Note that all these experiments are run separately for Places2 and CelebA. The performance of the 3 detection networks are summarized in Table~\ref{ijcai2021:tab0}.

\paragraph{Overall performance.}
As shown in Table~\ref{ijcai2021:tab0}, our NIX-Net outperforms existing methods by a large margin in all test scenarios, especially when transferred to detect unseen inpainting methods. When trained on the hybrid datasets, our NIX-Net achieved the best overall performance. Next, we will provide a detailed analysis of these results from two perspectives: 1) the effectiveness of different detection networks, and 2) the importance of universal training data.

\paragraph{Effectiveness of different detection networks.}
For LDICN and ManTra-Net, although decent results can be obtained on the known (used for generating training data) inpainting method, their performance drops drastically on unseen inpainting methods.
Such a poor generalizability indicates that both models overfit to the artifacts of a particular inpainting method and fail to consider the common characteristics of different deep inpainting techniques.
By contrast, our NIX-Net demonstrates consistently better generalizability, regardless of the inpainting method used for training. This is largely due to the sufficient (multi-scale and cross fusion) exploitation of the noise information contained in real versus inpainted contents.
This also indicates that noise patterns are indeed a reliable cue of detecting inpainted regions.
% The reason is that NIX-Net can effectively exploit the noise information and incorporate it into the detection procedure.

\paragraph{Importance of universal training data.}
Revisit Table~\ref{ijcai2021:tab0}, we find that, whenever the UT dataset is used in conjunction with one inpainting-method-aware dataset, the generalization performance of our NIX-Net can be significantly improved. Moreover, the UT dataset alone can lead to much better generalizability of existing methods LDICN and ManTra-Net. This result verifies, from the data perspective, the importance of noise modeling for universal deep inpainting detection. More importantly, such noise modeling like our proposed universal training dataset generation is much easier than generating training data using different deep inpainting techniques, making it more practical for real-world applications.
Another important observation is that NIX-Net trained on UT alone can achieve a similar level of performance as it was trained on the hybrid datasets, though combining more training data does improve the performance.
% significant improvement on generalization ability has been achieved when refining our NIX-Net with the proposed Universal Training (UT) data as additional training data.
% This is because of that UT data imitates the noise inconsistency of deep inpainting and thus alleviates the overfitting to approach-specific artifacts.

% Moreover, we present the results of the three models using only UT data as the training set.
% All the models obtain decent generalization ability and our model still shows overall better results on the test sets.
% This verifies that, even without training images yielded by any of the deep inpainting approach, our proposed universal training data is capable of realizing competitive detection accuracy and can be applied on different detection models. 
% Besides, it also shows the superiority of NIX-Net compared to the other two detection models.

\subsection{Qualitative Performance Evaluation}
Here, we provide a qualitative comparison by visualizing the detected masks. Figure~\ref{ijcai2021:QualitativeCompare1} illustrates some of the examples on Places2 and CelebA for LDICN/ManTra-Net trained only on data generated by GL and our NIX-Net trained only on UT data. The ten tested images are all inpainted by CA.
% Both LDICN and ManTra-Net are only trained with data generated by GL.
By comparing the ground truth mask (Mask GT) and the masks predicted by different detection networks, one can find that our NIX-Net can produce the most similar masks to the ground truth. 
LDICN and ManTra-Net, however, cannot accurately identify the inpainted regions, especially when they are complex (the three bottom rows). These visual inspections confirm the superiority of our proposed universal training data generation approach and the NIX-Net detection network.
% This confirms that the two baselines suffer from limited generalization ability.
% In contrast, our model is only trained with the proposed universal training data.
% It can be clearly seen from the figure that the detected masks by our model better capture the shape of the corresponding ground-truth masks.
% We thus claim that our method can provide more univeral and reliable detection for unseen deep inpainting approaches. 

\subsection{Ablation Study}
Here, we run a set of ablation studies to provide a complete understanding of the two key components of our NIX-Net network: two-steam (noise+image) feature learning and multi-scale cross fusion. 
Table~\ref{ijcai2021:tab2} compares the full NIX-Net detection network with its five variants created by removing or keeping the noise/image stream or the three fusion modules.
All these networks are trained on the UT dataset generated from Places2 and tested on test images from Places2 by GL, CA and GC. 
It shows that, after removing either the noise or the image stream, the performance degrades drastically. This implies that both the image and the noise pattern are crucial for extracting rich features for detection. The worst performance is observed when all 3 fusion modules removed, even though it still has the noise and the image streams.
When adding either fusion module 1 and 2 or fusion module 3 back into the network, the performance is clearly improved.
These results indicate that the proposed fusion module is essential for exchanging the information across multi-scale representations and achieving semantically richer feature fusion.

%(a) "w/o noise branch", which removes the noise stream in the feature extraction component; (b) "w/o multi-scale", which only takes the feature maps of image and noise streams at $1/8$ scale, and fuses them by direct concatenation on the channel dimension.
%It is obvious that both strategies are crucial and without either one will degrade the performance.

\begin{table}[thb]
\renewcommand\arraystretch{1.05}
\begin{center}
\scalebox{0.85}
{
\begin{tabular}{l|ccc}
\toprule
 \multirow{2}{*}{\textbf{Ablation of NIX-Net}} & \multicolumn{3}{c}{\textbf{Test mIoU}} \\ \cline{2-4}
& GL & CA & GC \\ \midrule
w/o noise stream & 88.24 & 84.11 & 79.87 \\ 
w/o image stream & 83.67 & 77.59 & 74.14 \\ 
\hline
w/o all fusion modules & 79.36  & 76.22 & 67.84 \\ 
w/o fusion module 1 and 2 & 84.72  & 82.35 & 77.93 \\ 
w/o fusion module 3 & 89.19  & 85.35 & 81.44 \\ 
\hline
Full NIX-Net &  \textbf{91.33} & \textbf{88.46}  &  \textbf{84.71}   \\ \bottomrule

\end{tabular}

}
\end{center}
\caption{Ablation of NIX-Net for (a) the noise/image stream or (2) the 3 multi-scale cross fusion modules. Networks are trained on the UT dataset generated for Places2 and tested on GL/CA/GC test sets.}
\label{ijcai2021:tab2}
\end{table}
\vspace{-0.1in}
%\subsection{Robustness to Noise}

\section{Conclusion}
In this work, we have proposed %a simple but very 
an effective approach for universal deep inpainting detection. %Specifically, we have introduced a new  universal training dataset generation method and a new  Noise-Image Cross-fusion (NIX-Net) detection network.
Our approach consists of two important designs: 1) a novel universal training dataset generation method and 2) a Noise-Image Cross-fusion (NIX-Net) detection network. %These two designs are motivate by the distinctive noise patterns exhibit in real versus generated contents.
Extensive experiments on two benchmark datasets verify the effectiveness of our proposed approach and its superior generalization ability when applied to detect unseen deep inpainting methods. Our work not only provides a powerful universal detection method but also opens up a new direction for building more advanced universal deep inpainting detectors.

\bibliographystyle{named}
\bibliography{ijcai21}

\end{document}